\documentclass[journal]{IEEEtran}
\usepackage{graphicx}	
\usepackage{url}
\usepackage{epsfig}
\usepackage{subcaption}
\usepackage{amssymb,amsmath}
\usepackage{multirow}

\usepackage[usenames]{color} 
\usepackage{booktabs}
\usepackage{enumerate}
\usepackage{color}
\usepackage{xcolor}
\usepackage{algorithm}
\usepackage{algpseudocode}
\usepackage{url}
\usepackage{comment}
\usepackage{placeins}
\usepackage{float}
\usepackage{enumerate}
\usepackage[hang,flushmargin]{footmisc}

\usepackage{balance}

\ifCLASSINFOpdf
\else
\fi
\begin{document}
%
\title{Multimodal Video-based Apparent Personality Recognition Using 
Long Short-Term Memory and \\ Convolutional Neural Networks}
%
%
%

\author{S\"{u}leyman~Aslan,,~\IEEEmembership{Student~Member,~IEEE,}
        and~U\u{g}ur G\"{u}d\"{u}kbay,~\IEEEmembership{Senior~Member,~IEEE,}
\thanks{The authors are with the Department
of Computer Engineering, Bilkent University, Ankara,
06800, Turkey, e-mail: (suleyman.aslan@bilkent.edu.tr, gudukbay@cs.bilkent.edu.tr).}
\thanks{Manuscript received Xxxxx XX, 20XX.}}%

%
%

\markboth{Journal of \LaTeX\ Class Files,~Vol.~14, No.~8, August~2015}%
{Shell \MakeLowercase{\textit{et al.}}: Bare Demo of IEEEtran.cls for IEEE Journals}
%



\maketitle

\begin{abstract}
Personality computing and affective computing, where the recognition of personality traits is essential, have gained increasing interest and attention in many research areas recently. We propose a novel approach to recognize the Big Five personality traits of people from videos. Personality and emotion affect the speaking style, facial expressions, body movements, and linguistic factors in social contexts, and they are affected by environmental elements. We develop a multimodal system to recognize apparent personality based on various modalities such as the face, environment, audio, and transcription features. We use modality-specific neural networks that learn to recognize the traits independently and we obtain a final prediction of apparent personality with a feature-level fusion of these networks. We employ pre-trained deep convolutional neural networks such as ResNet and VGGish networks to extract high-level features and Long Short-Term Memory networks to integrate temporal information. We train the large model consisting of modality-specific subnetworks using a two-stage training process. We first train the subnetworks separately and then fine-tune the overall model using these trained networks. We evaluate the proposed method using ChaLearn First Impressions V2 challenge dataset. Our approach obtains the best overall ``mean accuracy'' score, averaged over five personality traits, compared to the state-of-the-art.

\end{abstract}

\begin{IEEEkeywords}
deep learning, Convolutional Neural Network (CNN), Recurrent Neural Network (RNN), Long Short-Term Memory (LSTM) network, personality traits, personality trait recognition, multimodal information.
\end{IEEEkeywords}

%
\IEEEpeerreviewmaketitle

\section{Introduction}
\label{section1}
%
%
%
%

\IEEEPARstart{P}{ersonality} and emotions have a strong influence on people's lives and they affect behaviors, cognitions, preferences, and decisions. Emotions have distinct roles in decision making, such as providing information about pleasure and pain, enabling rapid choices under time pressure, focusing attention on relevant aspects of a problem, and generating commitment concerning decisions~\cite{pfister2008multiplicity}. Additionally, research suggests that the human decision-making process can be modeled as a two systems model, consisting of rational and emotional systems~\cite{kahneman2011thinking}. Accordingly, emotions are part of every decision-making process instead of simply having an effect on these processes. Likewise, personality also has an important effect on decision making and it causes individual differences in people's thoughts, feelings, and motivations. It can be observed that there are significant relationships among attachment styles, decision-making styles, and personality traits~\cite{deniz2011investigation}. In addition, personality relates to individual differences in preferences, such as the use of music in everyday life~\cite{chamorro2007personality,rentfrow2003re}, and user preferences in multiple entertainment domains including books, movies, and TV shows~\cite{cantador2013relating}. Due to the fact that emotion and personality have an essential role in human cognition and perception, there has been a growing interest in recognizing the human personality and affect and integrating them into computing to develop artificial emotional intelligence, which is also known as ``affective computing''~\cite{picard2000affective}, in combination with ``personality computing''~\cite{vinciarelli2014survey}. Hence, it becomes essential to recognize the personality and emotion of humans precisely. We present a novel multimodal framework to recognize the apparent personality of individuals from videos to address this problem.

\subsection{Personality Traits}

Personality can be defined as the psychological factors that influence an individual's patterns of behaving, thinking, and feeling that differentiate the individual from one another~\cite{cervone2013personality,mccrae1992introduction}. The most mainstream and widely accepted framework for personality among psychology researchers is the Five-Factor Model (FFM)~\cite{mccrae1992introduction,McCrae1987ValidationOT}. FFM is a model based on descriptors of human personality along five dimensions as a complete description of personality. Various researchers identified the same five factors within independent works in personality theory~\cite{McCrae1987ValidationOT,TupesRecurrentPersonality,GoldbergPhenotypicPersonality}. Therefore, it is considered reliable to define personality with FFM.

Based on the work by Costa, McCrae, and John~\cite{mccrae1992introduction,McCrae1987ValidationOT}, the five factors are defined as follows.

\begin{itemize}
	\item \textit{Openness (O):} Appreciation of experience and curiosity of the unfamiliar.
	\item \textit{Conscientiousness (C):} Level of organization and being dependable.
	\item \textit{Extraversion (E):} Social activity and interpersonal interaction.
	\item \textit{Agreeableness (A):} Tendency to work cooperatively with others and avoiding conflicts.
	\item \textit{Neuroticism (N):} Emotional instability and being prone to psychological distress.
	
\end{itemize}

\begin{table}[ht]
	\centering
	\caption{The characteristics of personality traits}
	\begin{tabular}{|l|l|l|} 
		\hline
		\textit{Low Scorer} & \textit{Personality Trait} & \textit{High Scorer} \\ \hline
		Calm, secure                        & Neuroticism        & Nervous, sensitive       \\ \hline
		Quiet, reserved                     & Extraversion       & Talkative, sociable      \\ \hline
		Cautious, conventional              & Openness           & Inventive, creative      \\ \hline
		Suspicious, uncooperative           & Agreeableness      & Helpful, friendly        \\ \hline
		Careless, negligent                 & Conscientiousness  & Organized, reliable      \\ \hline
	\end{tabular}
	\label{table:traitscale}
\end{table}

These five factors lead to bipolar characteristics that can be seen in individuals that score low and high on each trait, as seen in Table~\ref{table:traitscale}. The factors are often represented by the acronym \textit{OCEAN}.

The proposed model predicts the traits based on various modalities including facial, ambient, audio, and transcription features. We make use of convolutional neural networks (CNNs) in combination with Long Short-Term Memory (LSTM) networks in a two-stage training phase. The proposed approach obtains state-of-the-art results using ChaLearn First Impressions V2 (CVPR'17) challenge dataset~\cite{ponce2016chalearn}. 

\subsection{Contributions}
The contributions to the area of automatic recognition of people's apparent personality from videos are as follows:

\begin{itemize}
	\item[i)] A multimodal neural network architecture to recognize apparent personality traits from various modalities such as the face, environment, audio, and transcription features. The system consists of modality-specific deep neural networks that aim to predict apparent traits independently where the overall prediction is obtained with a fusion.
	\item[ii)] Integrating the temporal information of the videos learned by LSTM networks to the extracted spatial features with CNNs such as facial expressions and ambient features.
	\item[iii)] A two-stage training method that trains the modality-specific networks separately in the first stage and fine-tunes the overall model to recognize the traits accurately in the second stage.
\end{itemize}

The rest of the paper is organized as follows. Section~\ref{section3} discusses the related work. Section~\ref{section4} presents the proposed method that effectively learns a mapping from multimodal data to apparent personality trait vectors. Section~\ref{section5} shows the results of the approach and evaluates it with different quality aspects. Finally, Section~\ref{section6} concludes the paper and provides possible future research directions.

\section{Personality Recognition}
\label{section3}

Personality computing benefits from methods aimed towards understanding, predicting, and synthesizing human behavior~\cite{vinciarelli2014survey}. The effectiveness in analyzing such important aspects of individuals is the main reason behind the growing interest in this topic. Automatic recognition of apparent personality is a part of many applications such as human-computer interaction, computer-based learning, automatic job interviews,  autonomous agents, and crowd simulations~\cite{ponce2016chalearn,nass2001does,tlili2016role,DurupinarPAGB11,Bera2017,Basak2018}. Similarly, emotion is incorporated into adaptive systems in order to improve the effectiveness of personalized content and bring the systems closer to the users~\cite{tkalvcivc2016introduction}. As a result, personality and emotion-based user information is used in many systems, such as affective e-learning~\cite{shen2009affective}, conversational agents~\cite{ball2000emotion}, crowd simulations~\cite{Durupinar2016}, and recommender systems~\cite{recio2009personality,hu2010study,arapakis2009integrating}. Overall, personality is usually relevant in any system involving human behavior. Rapid advances in personality computing and affective computing led to the releases of novel datasets for apparent personality and emotional states of people from various sources of information such as physiological responses or video blogs~\cite{ponce2016chalearn,Subramanian2018}. One of the latest problems is recognizing five personality traits automatically from videos of people speaking in front of a camera. 

\subsection{Related Work}

Recently, there have been many approaches to recognizing personality traits. By analyzing the audio from spoken conversations~\cite{valente2012annotation} and based on the tune and rhythm aspects of speech~\cite{madzlan2014} it is possible to annotate and recognize the personality traits or predict the speaker attitudes automatically. These approaches demonstrate that audio information is important for personality. 

The status text of users on social networks can be utilized for the recognition of personality traits~\cite{alam2013personality} and it is also possible to explore the projection of personality, especially extraversion, through specific linguistic factors across different social contexts using transcribed video blogs and dialogues~\cite{nowson2014look}. Therefore, it is indicated that there is a strong correlation between users' behavior on social networks and their personality~\cite{Farnadi2016}. 

There are methods for recognition based on combinations of speaking style and body movements. Personality traits can be automatically detected in social interactions from acoustic features encoding specific aspects of the interaction and visual features such as head, body, and hands fidgeting~\cite{pianesi2008multimodal}. Likewise, five-factor personality traits can be automatically detected in short self-presentations based on the effectiveness of acoustic and visual non-verbal features such as pitch, acoustic intensity, hand movement, head orientation, posture, mouth fidgeting, and eye-gaze~\cite{batrinca2011please}. These features can be extracted from multimodal data in human-machine and human-human interaction scenarios~\cite{Batrinca16}. Consequently, body gestures, head movements, facial expressions, and speech based on naturally occurring human affective behavior leads to effective assessment of personality and emotion~\cite{zeng2009survey}.  

Facial physical attributes from ambient face photographs can be an important factor in modeling trait factor dimensions underlying social traits~\cite{vernon2014modeling}. It can be seen that valid inferences for personality traits can be made from the facial attributes. This is supported by experiments that are carried out in order to evaluate personality traits and intelligence from facial morphological features~\cite{qin2018modern}, to predict the personality impressions for a given video depicting a face~\cite{gurpinar2016combining}, and to identify the personality traits from a face image~\cite{fernando2016persons}. A study investigating why CNNs are very successful for automatically recognizing personality traits of people speaking to a camera shows that face provides the most discriminative information for this task and CNNs primarily inspect crucial parts of the face, such as mouth, nose, and eyes~\cite{Ventura17}.

Some studies support the idea that it is human behavior to evaluate individuals by their faces with respect to their personality traits and intelligence since self-reported personality traits can be predicted reliably from a facial image~\cite{qin2018modern}, and impressions that influence people's behavior towards other individuals can be accurately predicted from videos~\cite{gurpinar2016combining}. Additionally, predicting personality factors for personality-based sentiment classification is shown to be beneficial in the analysis of public sentiment implied in user-generated content~\cite{lin2017personality}. Because the personality has an effect on various modalities, automatic recognition of personality traits is accomplished by combining multiple features in different studies to exploit a multimodal approach~\cite{pianesi2008multimodal,gurpinar2016combining,ZhangECCV2016,Wei2018}. 

According to~\cite{sarkar2014feature}, attributes and features such as audio-visual, text, demographic and sentiment features are essential parts of a personality recognition system. Although multimodal approaches are commonly used to recognize personality traits, relatively limited work has been done to develop a comprehensive method utilizing a considerable amount of informative features. In this work, we propose such a comprehensive method to recognize personality trait factors.

\section{The Proposed Framework}
\label{section4}

In our framework, we take a video clip of a single person as input and predict the personality traits associated with that person. The proposed framework is based on learning personality features separately using different modality-specific neural networks, then combining those learned high-level features to obtain a final prediction of personality traits. For this purpose, four neural networks are trained independently to extract high-level features, namely, \textit{ambient features}, \textit{facial features}, \textit{audio features}, and \textit{transcription features}.

\begin{figure}[ht]
	\centering
	\fbox{\includegraphics[width=0.85\columnwidth]{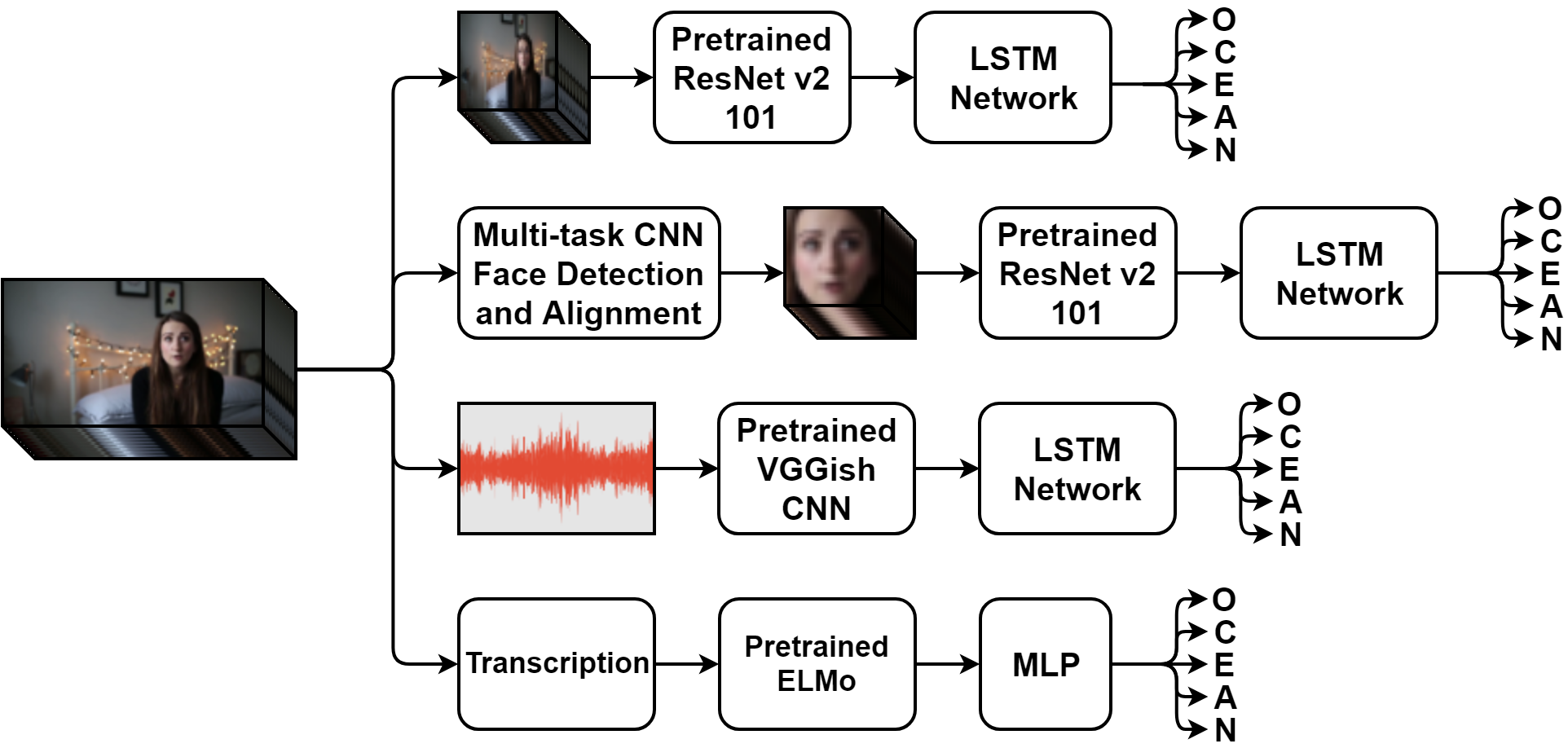}}
	\caption{The first stage of the proposed model. Subnetworks learn to recognize personality traits based on the corresponding input features.}
	\label{fig:modelfirststage}
\end{figure}

\begin{figure}[ht]
	\centering
	\fbox{\includegraphics[width=0.984\columnwidth]{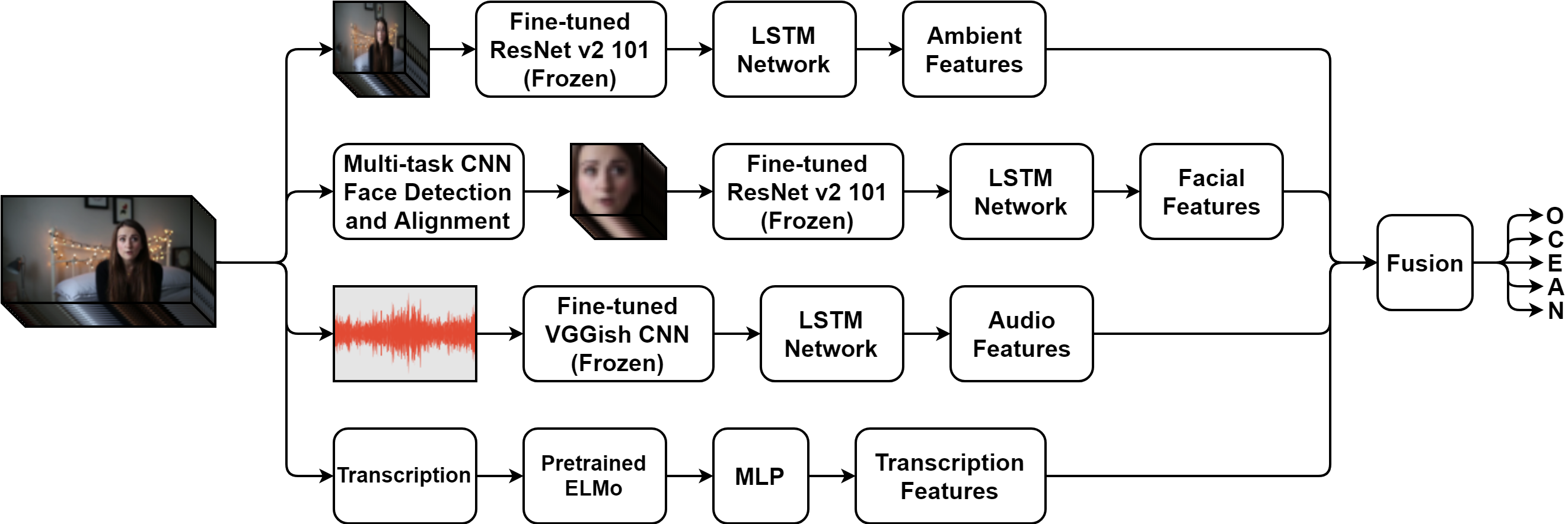}}
	\caption{The second stage of the proposed model. We use trained subnetworks as feature extractors and fuse the results of them to obtain the final score for traits.}
	\label{fig:modelsecondstage}
\end{figure}

In this approach, there are two stages. In the first stage, each subnetwork is trained to obtain personality traits according to the various input features. The flowchart illustrating this first stage is given in Figure~\ref{fig:modelfirststage}. Modality-specific networks are trained separately because first, to make sure that each network is able to learn corresponding features and improves the final prediction, and second, to prevent the model to focus on only one dominant feature in training phase. In the second stage, trained neural networks are used as feature extractors and the final trait scores are obtained through a fusion. The flowchart of second stage is given in Figure~\ref{fig:modelsecondstage}. We elaborate on each subnetwork in the following sections.

\subsection{Ambient Feature-based Recognition}

One of the approaches used in the proposed framework is to recognize personality traits based on the ambient features related to the person such as surrounding objects, lighting, and clothing. The intuition behind this approach is that those features can influence the apparent personality of the person. It has been demonstrated that environmental elements such as surroundings, colors, and lighting have an effect on the mood and perception~\cite{kasmar1968effect}. Additionally, these features provide more information about the video clip, which makes a deep neural network to learn more effectively.

We first sample the frames at equal intervals of one second because video clips can have varying frame rates and taking consecutive frames would be inconsistent in terms of the temporal relation of frames. Besides, for a video with a high frame rate and long duration, the total number of frames become quite large so training the neural networks would be unnecessarily slow and memory intensive. With uniform sampling, the learning process is efficient without losing significant information. Another preprocessing operation applied to the frames is resizing. Currently, images with high resolution such as frames of 720p or 1080p videos make training a CNN infeasible. Because of this reason, all frames are resized to 224$\times$224~pixels. Color information is retained and all of the frames have an RGB color space.

To recognize apparent personality trait factors from preprocessed frames, first, we use a CNN. Deep Convolutional Neural Networks (DCNN) achieve superior recognition results on a wide range of computer vision problems~\cite{goodfellow2014generative},~\cite{lecun2015deep} and they are most suitable for this task as well. We evaluate various deep neural networks and provide comparisons. We use a ``warm start'' of pretrained ResNet-v2-101~\cite{he2016identity}, trained on ILSVRC-2012-CLS image classification dataset~\cite{ILSVRC15}, with fine-tuning to fit the model to the problem. ResNet is a part of a larger model, where it corresponds to the lower layers of the model and more layers are added on top of ResNet.

\begin{figure}[ht]
	\centering
	\fbox{\includegraphics[width=0.984\columnwidth]{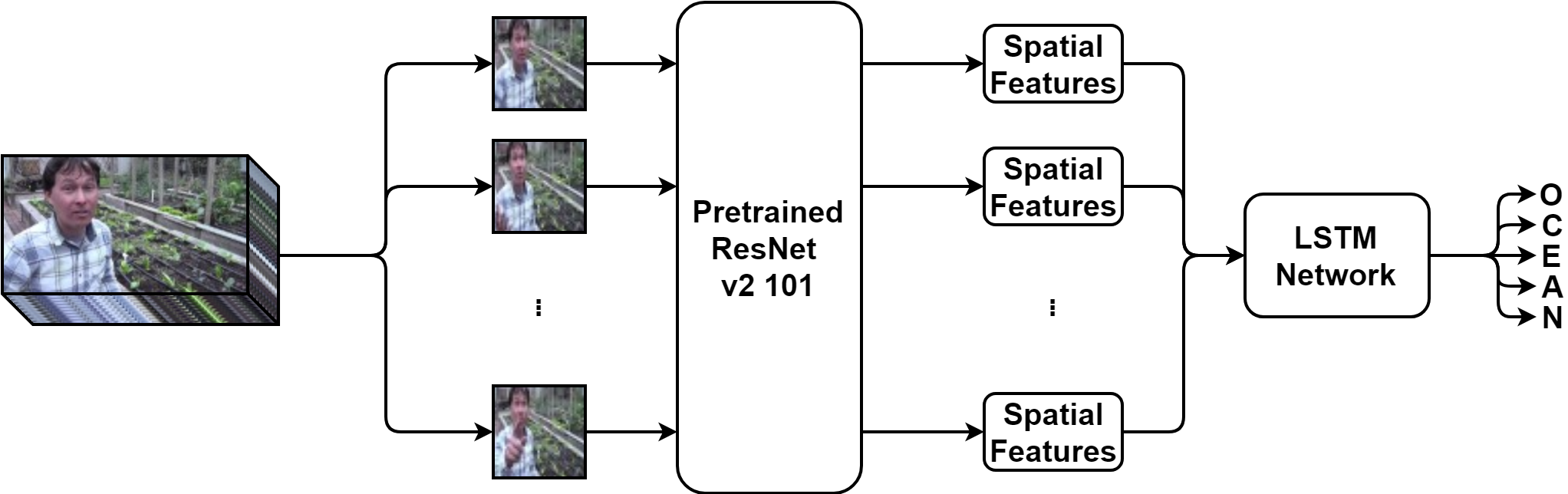}}
	\caption{The ambient feature-based neural network.}
	\label{fig:modelambient}
\end{figure}

We apply the CNN to the video on a per-frame basis so that high-level spatial features are learned. We then exploit the temporal information between video frames. Recurrent Neural Networks (RNNs) allow information from previous events to persist and can connect that information to the present event, however, it has been shown that RNNs are unable to learn dependencies that are long-term~\cite{hochreiter1991untersuchungen},~\cite{bengio1994learning}. LSTM networks are designed to address this problem and have been demonstrated to be successful~\cite{hochreiter1997long}. In order to integrate the temporal information, we experiment with RNNs and various types of LSTM networks, similar to CNNs. We use LSTM units based on Gers et al. style of LSTM networks~\cite{gers1999learning}. In this architecture, the LSTM network corresponds to the higher layers of the model. Figure~\ref{fig:modelambient} shows the architecture of this subnetwork.

We train this neural network to recognize personality traits based on ambient features only. Afterward, we use all of the trained layers in a larger model where the subnetwork will be a component of the model.

\subsection{Facial Feature-based Recognition}

One other approach in the proposed method is to recognize the traits based on facial features. It has been shown that personality can be accurately assessed from faces~\cite{little2007using} and facial symmetry is associated with five-factor personality factors~\cite{fink2005facial}. Hence, it is crucial to make use of this information in order to assess personality. Although faces are included in the images used by the previously mentioned subnetwork, they become too small after scaling and in order to analyze faces properly, other parts of the images should be removed. We use faces as the sole input of the neural network in this approach.

In order to obtain facial features, we first detect faces and align them. One face detector that has shown to work well is Multi-task CNN (MTCNN)~\cite{zhang2016joint} and we make use of MTCNN in our method. We apply face alignment to the frames used in the ambient feature-based subnetwork, to ensure that the time step is consistent across different modalities. Likewise, we scale frames after face alignment, resulting in 224$\times$224~pixel face images.

Because both facial features-based recognition and ambient features-based recognition are computer vision tasks, the rest of this process is similar. First, there is a CNN that learns high-level spatial features per-frame basis, then a recurrent neural network, specifically an LSTM network, integrates the temporal information. We use ResNet-v2-101~\cite{he2016identity} as the CNN because it is the most effective one according to the experiments. 

\begin{figure}[ht]
	\centering
	\fbox{\includegraphics[width=0.984\columnwidth]{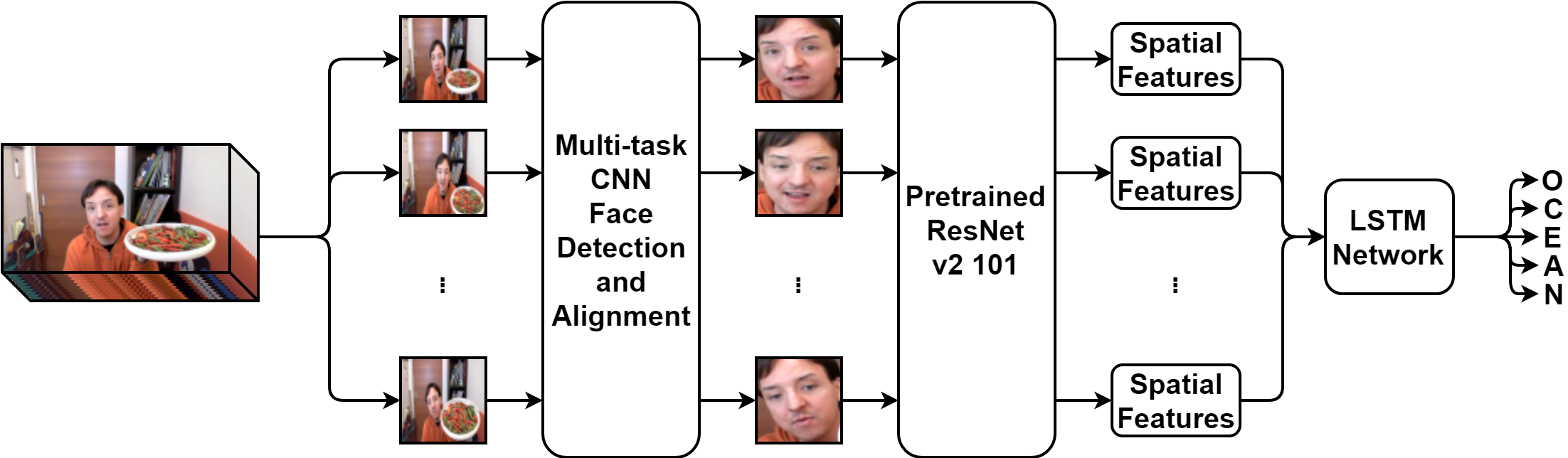}}
	\caption{The facial feature-based neural network.}
	\label{fig:modelfacial}
\end{figure}

The LSTM network built on top of the CNN is similar to the one for the ambient feature-based subnetwork. We train this neural network consisting of ResNet and LSTM units to learn features from aligned faces and to assess the five personality factors. Figure~\ref{fig:modelfacial} depicts the architecture of this subnetwork.

\subsection{Audio Feature-based Recognition}

The third modality used in the proposed model is the audio. We extract input features from the audio waveforms for the model before using a neural network. This process is the same as the preprocessing method used to train a VGG-like audio classification model, called VGGish~\cite{hershey2017cnn}, on a large YouTube dataset that is a preliminary version of YouTube-8M~\cite{abu2016youtube}. As a result of this process, we compute a log mel-scale spectrogram and convert these features into a sequence of successive non-overlapping patches of approximately one second for each audio waveform~\cite{Min2005}.

After we obtain the audio feature patches, we use a CNN to convert these features into high-level embeddings. The input of CNN is 2D log mel-scale spectrogram patches where the two dimensions represent frequency bands and frames in the input patch. Although we tested different architectures for the neural network, we used the pretrained VGGish model~\cite{hershey2017cnn} as a ``warm start'' and fine-tuned it in our framework. This model outputs 128-dimensional embeddings for each log mel-scale spectrogram patch.

\begin{figure}[ht]
	\centering
	\fbox{\includegraphics[width=0.984\columnwidth]{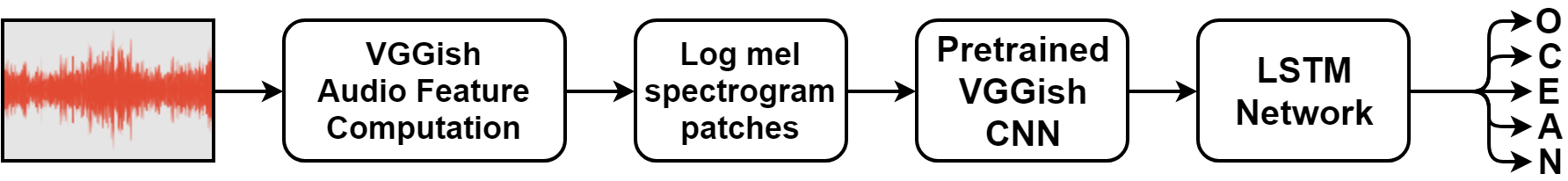}}
	\caption{The audio feature-based neural network.}
	\label{fig:modelaudio}
\end{figure}

Because there is a patch for each one-second interval and embeddings are obtained from these patches, we make use of temporal correlation before predicting the apparent personality. This part is the same as integrating the time information in the video for ambient feature-based and facial feature-based subnetworks; we use an LSTM network again. Consequently, this composition of neural networks learns to recognize personality from audio. Figure~\ref{fig:modelaudio} shows the architecture of this subnetwork.

\subsection{Transcription Feature-based Recognition}

The last modality used in the proposed method is the transcription of the speech of people in the videos. Psychological research has shown that personality influences the way a person writes or talks and word use and expressions are associated with personality~\cite{stemmler2010personality}. For example, individuals that score high in extraversion prefer complex, long writings and conscientious people tend to talk more about achievements and work~\cite{mehl2006personality}. These studies indicate that people with similar personality factors are likely to use the same words and choose similar sentiment expressions. Therefore, it is essential that this information is analyzed to make an accurate prediction of personality traits.

In this approach, we apply a language module to the text features in order to compute contextualized word representations and to encode the text into high dimensional vectors before the learning phase. For this purpose, there are several language models that can be applied. One particular approach that is suitable for this subnetwork is a language module that computes contextualized word representations using deep bidirectional LSTM units, which is trained on one billion word benchmark~\cite{chelba2013one}, called Embeddings from Language Models (ELMo)~\cite{peters2018deep}. This model outputs 1024-dimensional vector containing a fixed mean-pooling of all contextualized word representations. Although the model has four trainable scalar weights, in this setting, we fix all parameters so there is no additional training for this language module.

\begin{figure}[ht]
	\centering
	\fbox{\includegraphics[width=0.984\columnwidth]{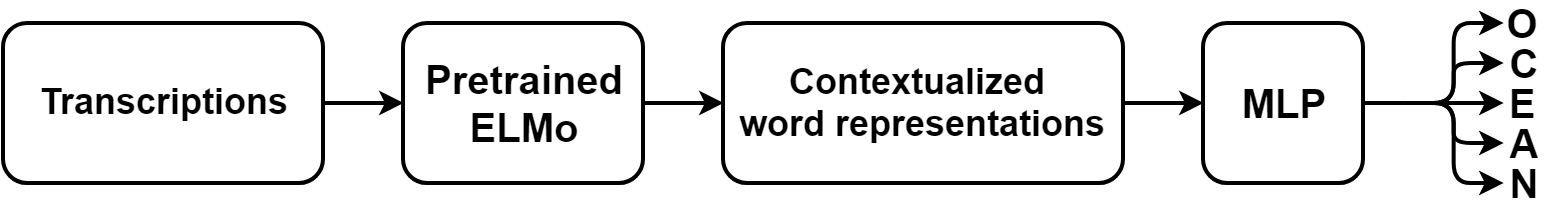}}
	\caption{The transcription feature-based neural network.}
	\label{fig:modeltranscription}
\end{figure}

After obtaining the embeddings, the next step is to directly learn to recognize personality traits from these features. At this stage, unlike all other subnetworks, there is no LSTM network or any other variation of RNNs because the information related to the sequences of words is already encoded into the embeddings through the bidirectional LSTM units in ELMo. As a result, a few additional layers on top of this language module are added to train a regressor neural network which performs recognition of personality factors from transcription features. Figure~\ref{fig:modeltranscription} depicts the architecture of this subnetwork.

\section{Experimental Results and Evaluation}
\label{section5}

In this section, we present the experiments that are carried out for the proposed method, the dataset which the model is trained on, and the experimental results. The proposed approach and various other alternatives are experimented with and compared to each other, and the best performing method is compared to the state-of-the-art. The results demonstrate that the proposed method outperforms the current state-of-the-art. 

\subsection{Dataset}

\begin{figure}[ht]
	\centering
	\begin{subfigure}[b]{0.49\columnwidth}
		\includegraphics[width=\columnwidth]{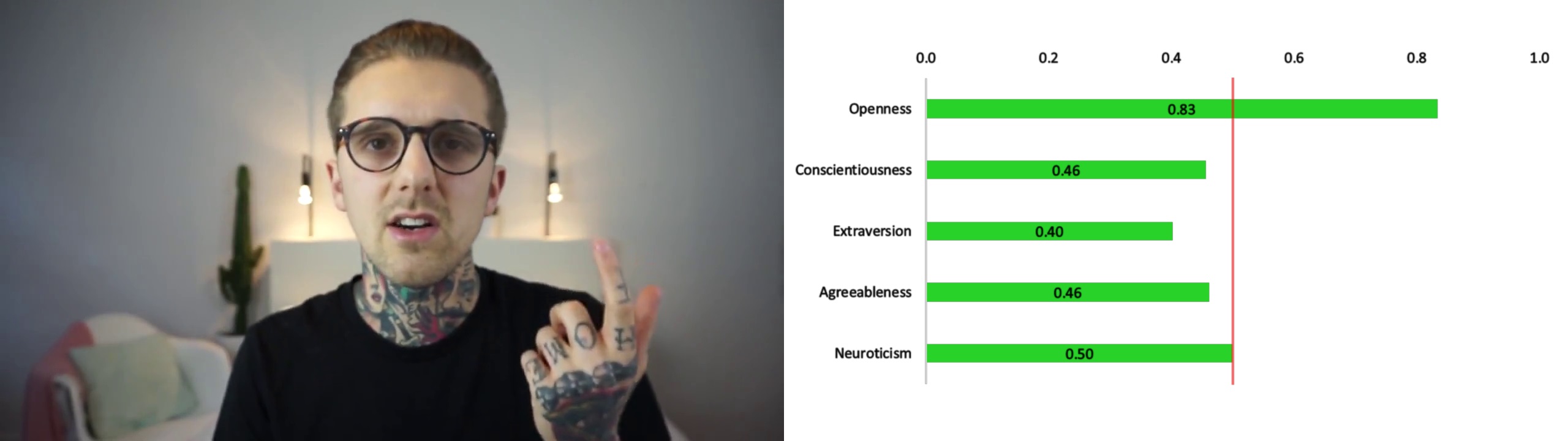}
	\end{subfigure}
	\begin{subfigure}[b]{0.49\columnwidth}
		\includegraphics[width=\columnwidth]{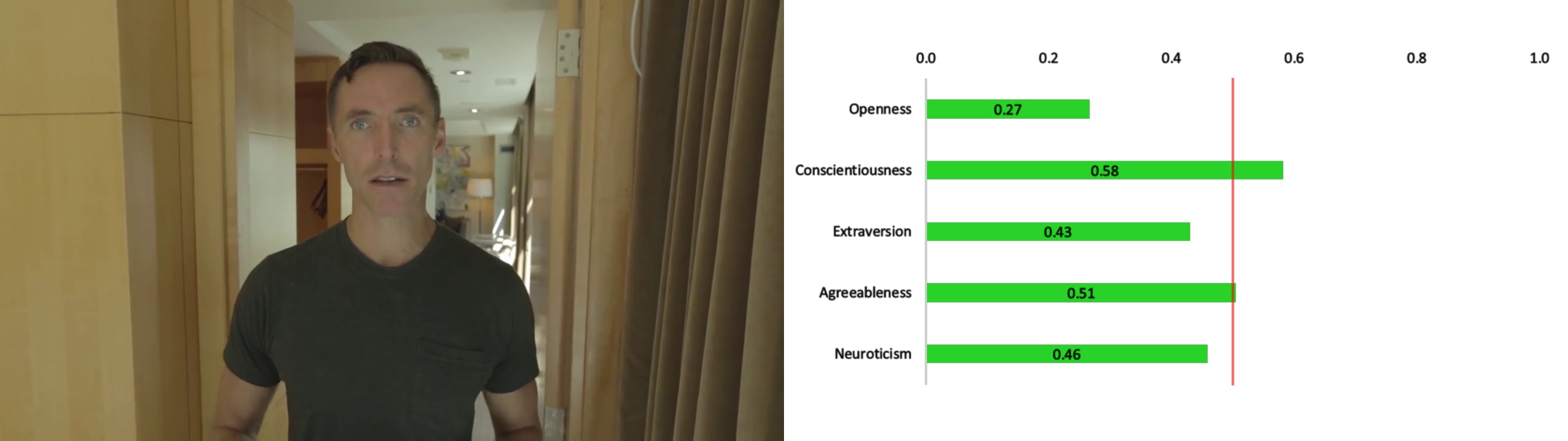}
	\end{subfigure}
	 \centerline{Openness}
	 \centerline{\ }
	\begin{subfigure}[b]{0.49\columnwidth}
		\includegraphics[width=\columnwidth]{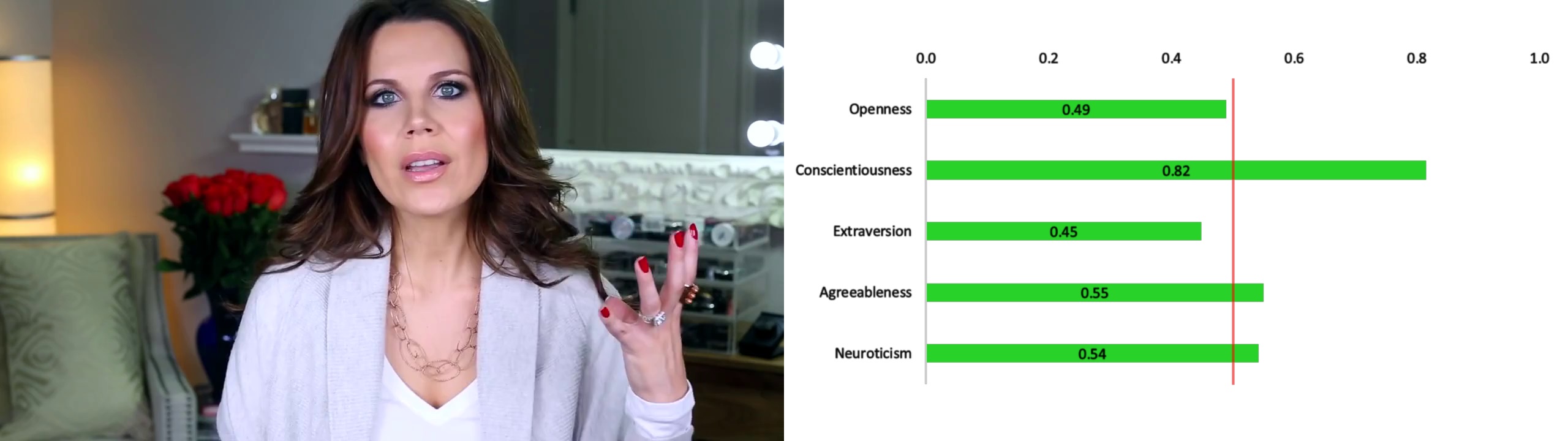}
	\end{subfigure}
	\begin{subfigure}[b]{0.49\columnwidth}
		\includegraphics[width=\columnwidth]{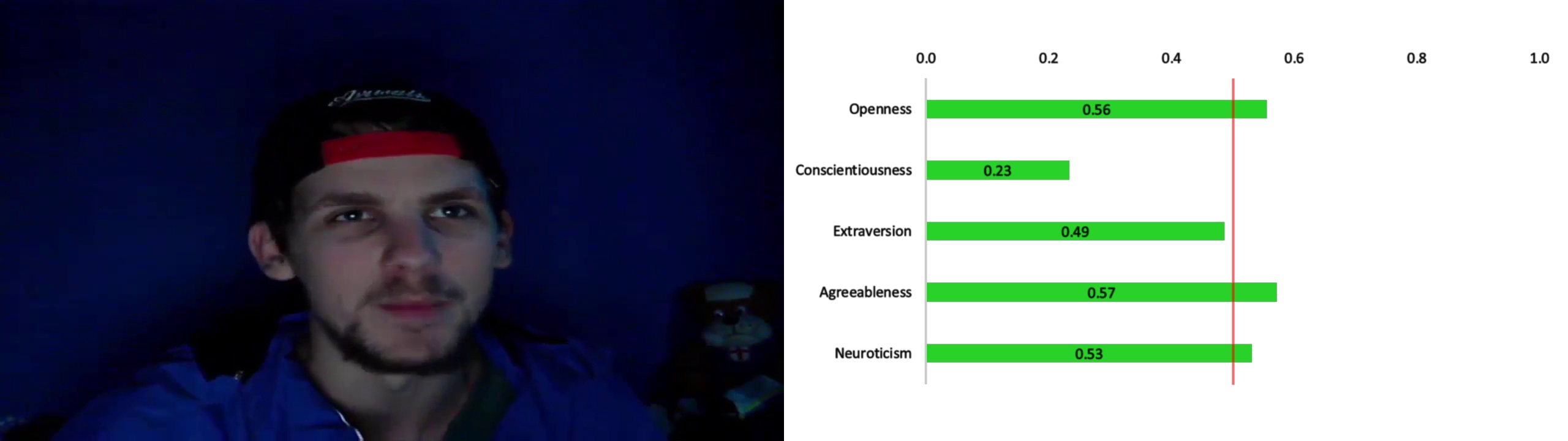}
	\end{subfigure}
	 \centerline{Conscientiousness}
 	 \centerline{\ }
	  \vspace*{2ex}
	\begin{subfigure}[b]{0.49\columnwidth}
		\includegraphics[width=\columnwidth]{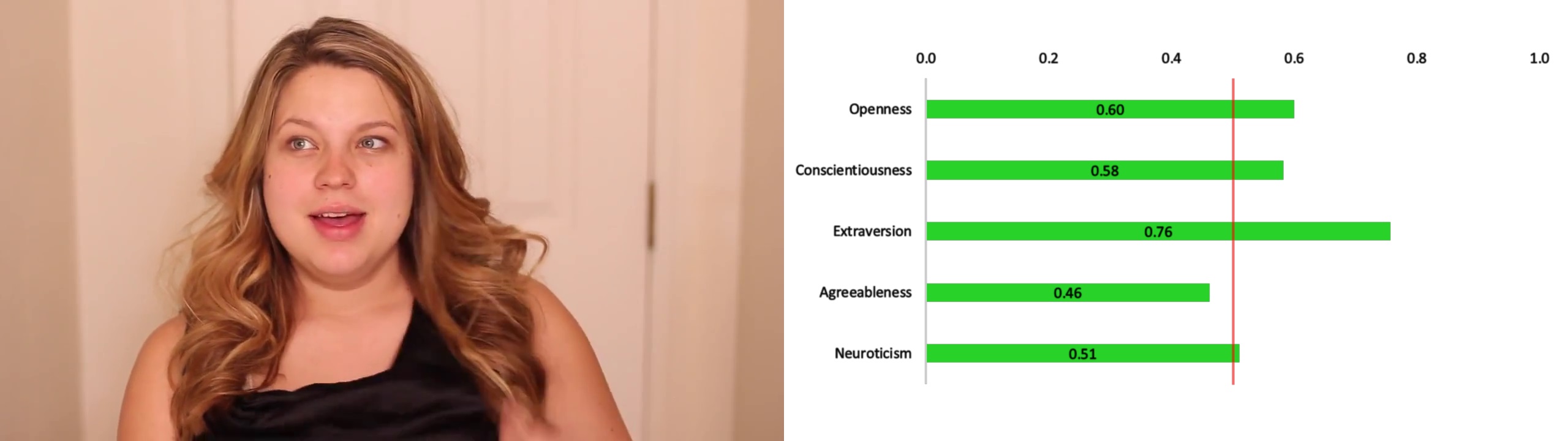}
	\end{subfigure}
	\begin{subfigure}[b]{0.49\columnwidth}
		\includegraphics[width=\columnwidth]{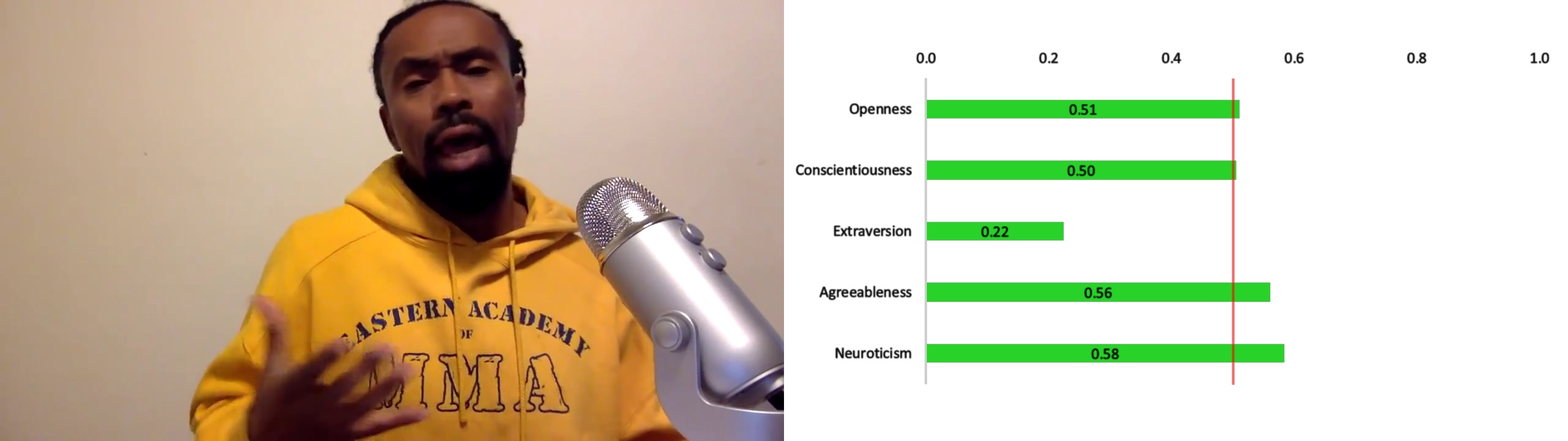}
	\end{subfigure}
	\centerline{Extraversion}
	 \centerline{\ }
	 \begin{subfigure}[b]{0.49\columnwidth}
		\includegraphics[width=\columnwidth]{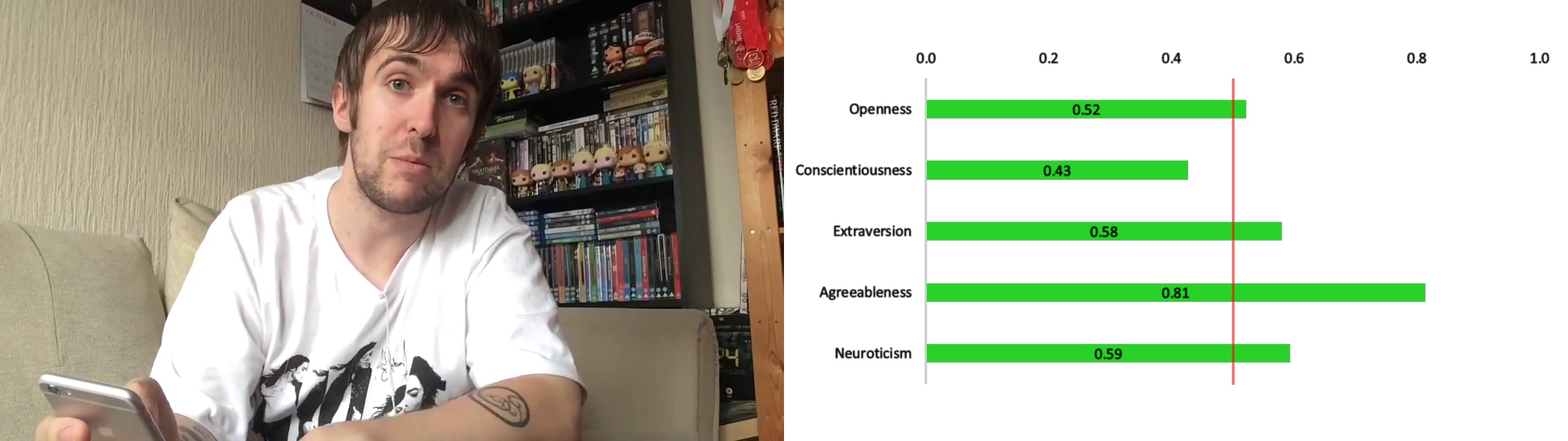}
	\end{subfigure}
	\begin{subfigure}[b]{0.49\columnwidth}
		\includegraphics[width=\columnwidth]{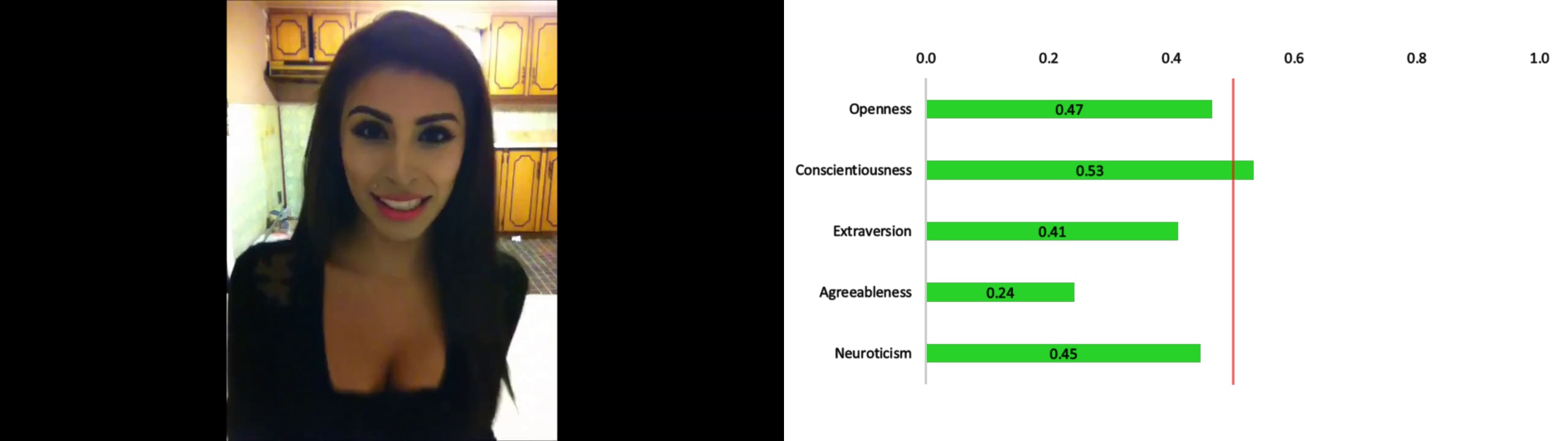}
	\end{subfigure}
	\centerline{Agreeableness}
 	 \centerline{\ }
	\begin{subfigure}[b]{0.49\columnwidth}
		\includegraphics[width=\columnwidth]{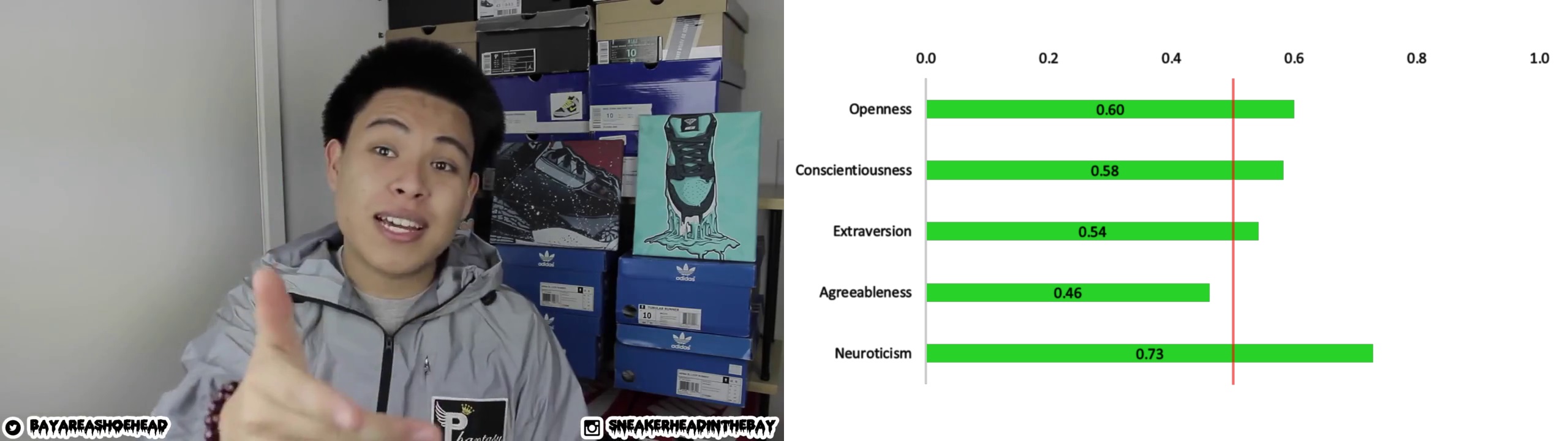}
	\end{subfigure}
	\begin{subfigure}[b]{0.49\columnwidth}
		\includegraphics[width=\columnwidth]{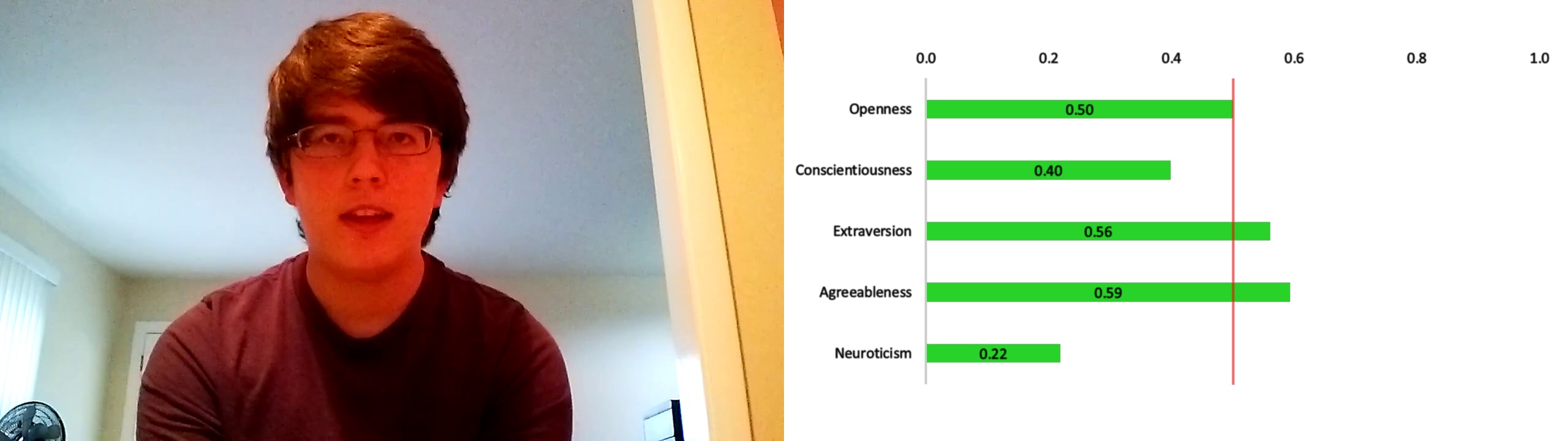}
	\end{subfigure}
	\centerline{Neuroticism}
	 \centerline{\ }
	\begin{subfigure}[b]{0.49\columnwidth}
		\includegraphics[width=\columnwidth]{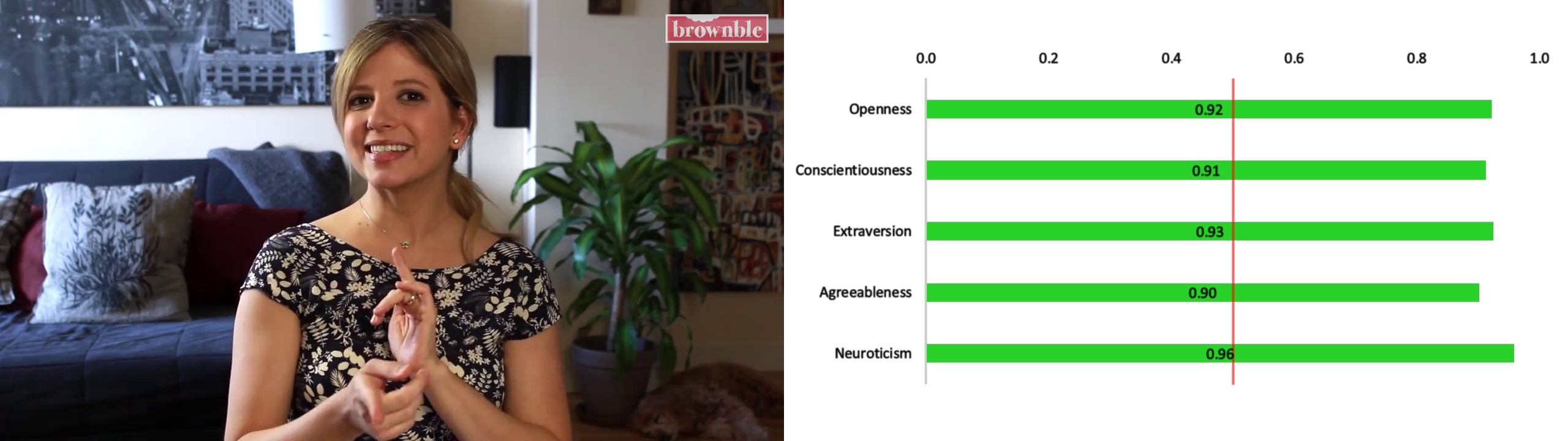}
	\end{subfigure}
	\begin{subfigure}[b]{0.49\columnwidth}
		\includegraphics[width=\columnwidth]{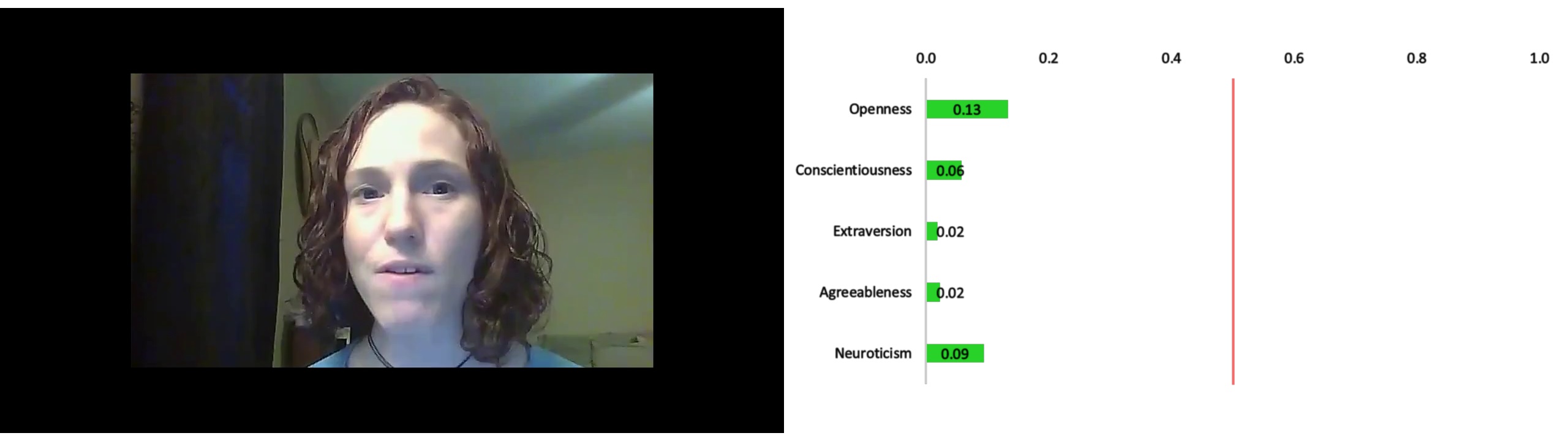}
	\end{subfigure}
	\centerline{All personality traits together}
	\caption{Sample videos from the training set depicting various cases of how personality traits are perceived by humans.}
	\label{fig:trainsetexamples}
\end{figure}

The dataset used to evaluate the proposed approach is the ChaLearn First Impressions V2 (CVPR'17) challenge dataset~\cite{ponce2016chalearn}, which is publicly available~\cite{ChaLearn}. The aim of this challenge is to automatically recognize apparent personality traits according to the five-factor model. The dataset for this challenge consists of 10000 videos of people facing and speaking to a camera. Videos are extracted from YouTube, they are mostly in high-definition (1280$\times$720~pixels), and in general, they have an average duration of 15 seconds with 30 frames per second. In the videos, people talk to the camera in a self-presentation context and there is a diversity in terms of age, ethnicity, gender, and nationality. The videos are labeled with personality factors using Amazon Mechanical Turk (AMT), so the ground truth values are obtained by using human judgment. For the challenge, videos are split into training, validation and test sets with a 3:1:1 ratio. Figure~\ref{fig:trainsetexamples} shows some examples of videos.

In the data collection process, AMT workers compared pairs of videos and evaluated the personality factors of people in the videos by choosing which person is likely to have more of an attribute than the other person for each personality factor~\cite{ponce2016chalearn}. Multiple votes per video, pairwise comparisons, and labeling small batches of videos are used to address the problem of bias for the labels. The final scores are obtained from the pairwise scores by using a Bradley-Terry-Luce (BTL) model~\cite{bradley1952rank}, while addressing the problem of calibration of workers and worker bias~\cite{chen2016overcoming}.

The evaluation metric is also defined by the challenge. From the trained models, it is expected that the model outputs continuous values for the target five personality traits in the range of [0, 1]. These values are produced separately for each trait, therefore there are 5 predicted values to be evaluated. For this purpose, the ``mean accuracy'' over all predicted personality trait values is computed as the evaluation metric~\cite{ponce2016chalearn}. Accordingly, it is defined as:

\begin{equation}
A = 1 - \frac{1}{N} \sum_{i=1}^{N} |t_i - p_i|
\end{equation}

\noindent where $t_i$ are the ground truth scores and $p_i$ are the estimated values for traits with the sum running over $N$ videos.

In the dataset, there are in total over 2.5 million frames that can be processed during training, and although encoded data size (size of videos) is about 27 gigabytes, decoded size (size of tensors) is over 10 terabytes. Because of this large-scale data, training CNNs and RNNs, which are computationally intensive, with the full usage of this dataset is infeasible for this task given the hardware used in experiments and time limitations. Therefore sampling and resizing are applied to this dataset prior to training the neural networks.

\subsection{First Stage Training}

The first stage consists of training each subnetwork independently and does not give a final prediction for personality traits. In this stage, we train the subnetworks as explained in the proposed framework as well as some alternatives to compare the results. We report the validation set performances of the subnetworks in Table~\ref{table:results_first_stage}. We optimize the hyperparameters for each system and in all systems we use Adam optimizer~\cite{kingma2014adam}. 

\begin{table}[htbp]
	\centering
	\caption{The validation set performance of the subnetworks.}
	\begin{tabular}{|l|c|} 
		\hline
		Subnetwork 	& Mean Accuracy \\ \hline
		Ambient: CNN & 0.9012 \\ \hline
		Ambient: 3D-CNN & 0.8962 \\ \hline
		Ambient: Inception-v2 & 0.9089 \\ \hline
		Ambient: ResNet-v2-101 & 0.9116 \\ \hline
		Face: MTCNN + Inception-v2 & 0.9067 \\ \hline
		Face: MTCNN + ResNet-v2-101 & 0.9136 \\ \hline
		Face: Dlib + Inception-v2 & 0.9058 \\ \hline
		Face: Dlib + ResNet-v2-101 & 0.9107 \\ \hline
		Audio: VGGish & 0.9049 \\ \hline
		Transcription: USE\_T & 0.8869 \\ \hline
		Transcription: ELMo & 0.8872 \\ \hline
		Transcription: Skip-gram & 0.8870 \\ \hline
	\end{tabular}
	\label{table:results_first_stage}
\end{table}

For visual feature-based subnetworks, we perform data augmentation during training, including adjusting the brightness, saturation, hue, and contrast of RGB images by random factors but not randomly flipping the images. We apply feature scaling to the input images as well because gradient descent converges much faster with feature scaling than without it~\cite{aksoy2001feature}. For the neural network architecture, we use Inception-v2~\cite{ioffe2015batch} and ResNet-v2-101~\cite{he2016identity} networks, both trained on ILSVRC-2012-CLS image classification dataset~\cite{ILSVRC15}, in addition to a simple CNN and a 3D-CNN. We remove the LSTM network from the system with a 3D-CNN because we capture temporal information through convolutions and there is no need for a recurrent neural network. In order to train the facial feature-based subnetwork, we perform face recognition and alignment using two methods, Dlib face detector~\cite{dlib09} and Multi-task CNN face detection and alignment~\cite{zhang2016joint}. 

For the transcription-based subnetwork, in addition to ELMo, we apply other language models to encode the transcription text into high dimensional vectors that can be used for personality recognition, such as Universal Sentence Encoder~\cite{cer2018universal} which is a model trained for natural language tasks such as semantic similarity, text classification, and clustering, and text embedding model based on the skip-gram version of word2vec~\cite{mikolov2013distributed,mikolov2013efficient}. To train the networks, we use a learning rate of $10^{-4}$ for audio feature-based subnetwork and $10^{-5}$ for other subnetworks, and a batch size of 8 videos with 6 frames for each video resulting in 48 images for the CNNs and 8 high-dimensional features from CNNs for the LSTM networks. The dropout probability is $0.5$.

According to the results, for the visual feature-based networks, we observe that ambient feature-based subnetwork obtains a ``mean accuracy'' score of $0.9116$ and the facial feature-based subnetwork outperforms that by obtaining a score of $0.9136$. Audio feature-based personality recognition is not as effective as vision feature-based recognition with a score of $0.9049$ and transcription feature-based network gives relatively worse performance by obtaining a score of only $0.8872$. The comparison of the subnetworks is given in Figure~\ref{fig:allsubnetworks}.

\subsection{Second Stage Training}

\begin{figure}[htbp]
	\centerline{
	\begin{subfigure}[b]{0.50\columnwidth}
		\includegraphics[width=\columnwidth]{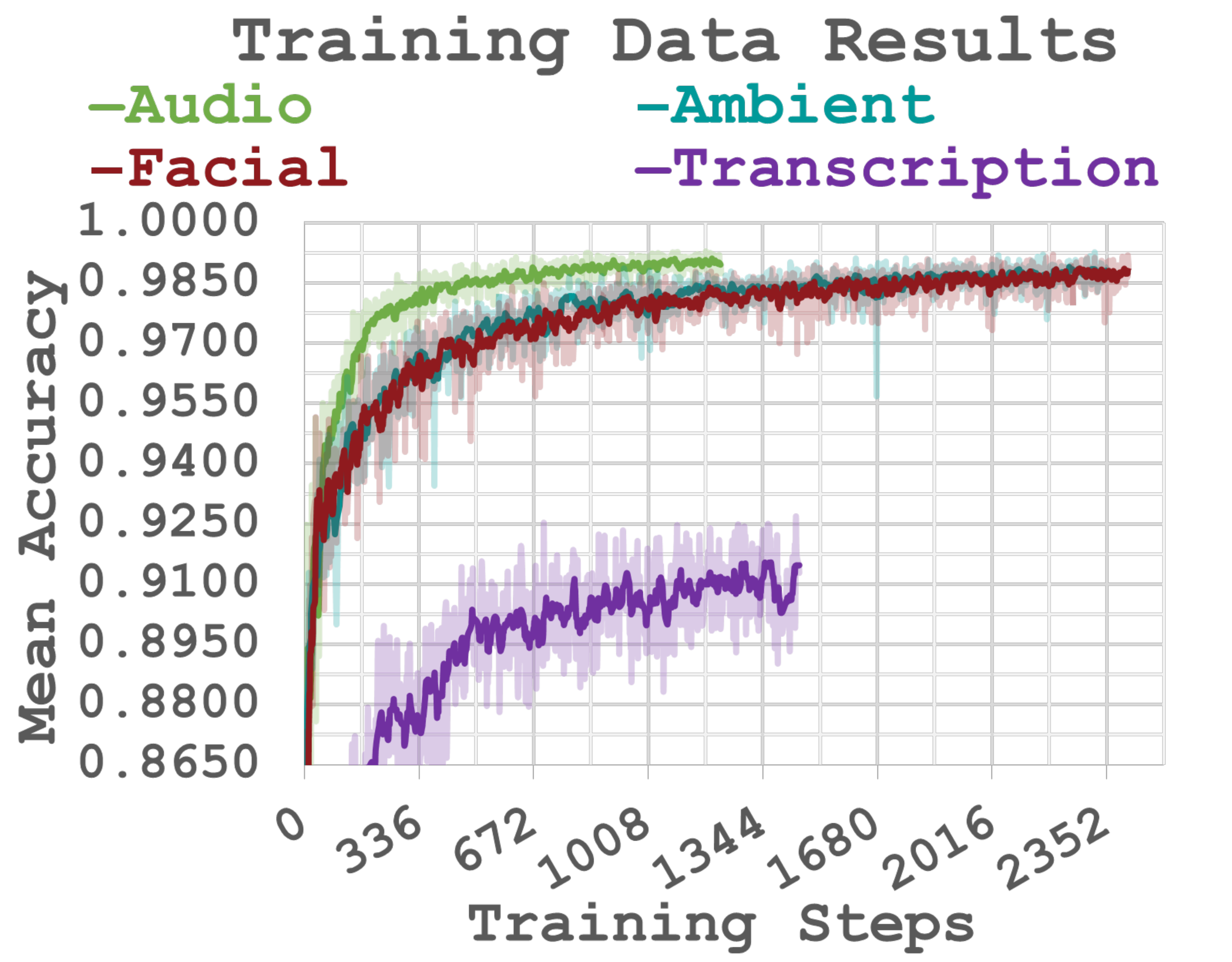}
	\end{subfigure}
	\begin{subfigure}[b]{0.50\columnwidth}
		\includegraphics[width=\columnwidth]{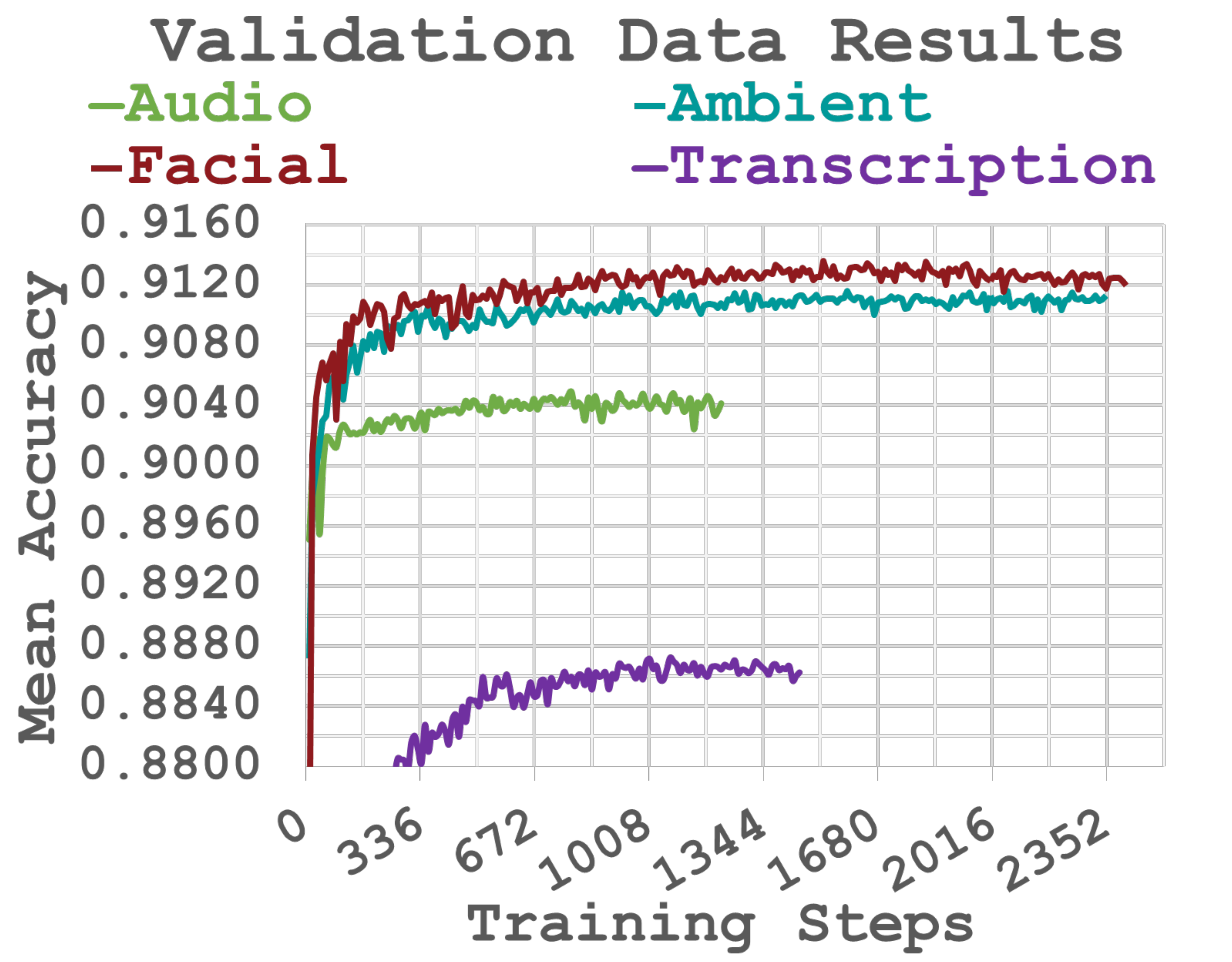}
	\end{subfigure}
	}
	\caption{The comparison of facial feature-based, ambient feature-based, audio feature-based, and transcription feature-based subnetworks.}
	\label{fig:allsubnetworks}
\end{figure}

The second stage of training the model consists of combining the separately trained modality-specific neural networks to obtain a final prediction of personality traits. For this purpose, we modify subnetworks in order to apply early feature-level fusion and train a larger model. We keep higher level features such as the outputs of ResNet-v2-101 and VGGish CNNs and the ELMo embeddings fixed while changing the outputs of each subnetwork by applying modifications to LSTM networks and dropping the last layer of the transcription feature-based subnetwork. The reasoning is that instead of getting five-dimensional outputs as the predicted personality traits from each subnetwork, we obtain higher-dimensional features so that the larger model can learn correlations between various modalities. Additionally, audio feature-based subnetwork has the tendency to overfit to the training data (see Figure~\ref{fig:allsubnetworks}). We reduce the capability of this subnetwork by changing network structures and increase the complexity of better performing subnetworks such as visual feature-based networks. For the evaluation, we combine modalities one by one in order to observe the effect of each subnetwork to the final model. To begin with, we use the ambient feature-based subnetwork and the audio feature-based subnetwork in order to utilize both visual and audio input. These subnetworks perform worse than facial feature-based subnetwork separately, so we compare the combined network to the facial feature-based subnetwork. 

\begin{figure}[htbp]
	\centerline{
	\begin{subfigure}[b]{0.50\columnwidth}
		\includegraphics[width=\columnwidth]{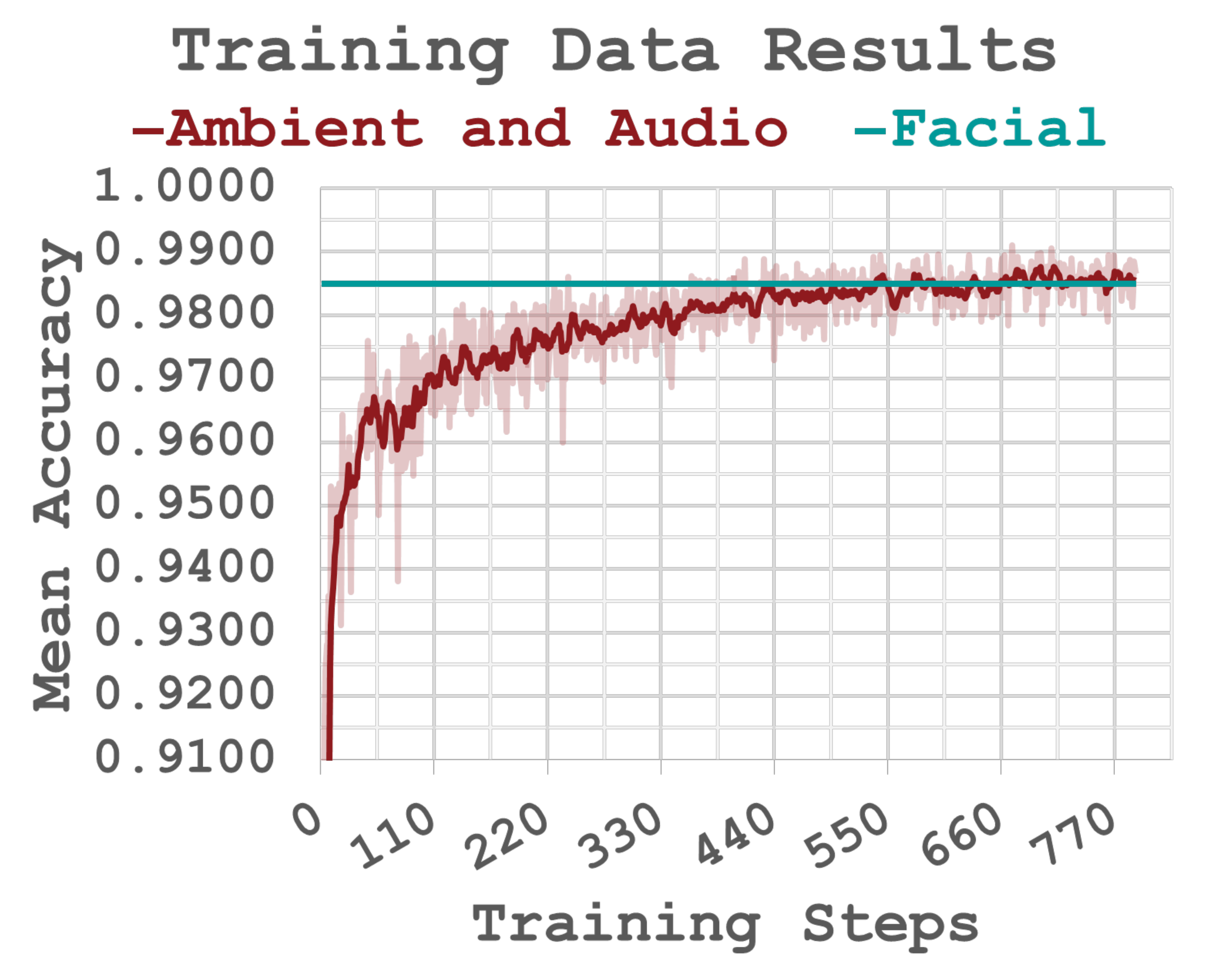}
	\end{subfigure}
	\begin{subfigure}[b]{0.50\columnwidth}
		\includegraphics[width=\columnwidth]{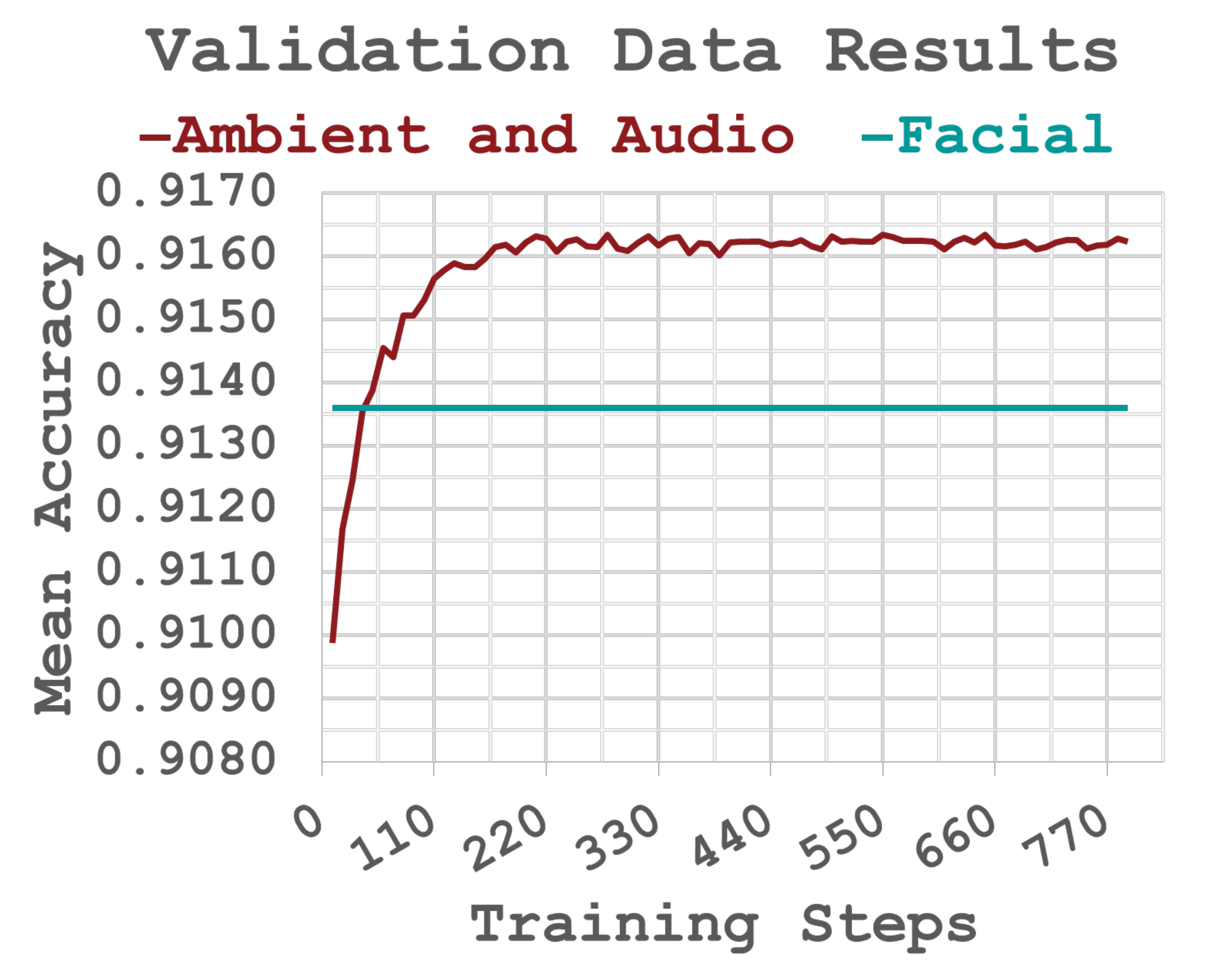}
	\end{subfigure}
	}
	\caption{The results of the simple multimodal network consisting of ambient feature-based and audio feature-based subnetworks with early fusion.}
	\label{fig:ambientaudio_vs_facial}
\end{figure}

According to the results, the score of the combined network is $0.9163$, outperforming the score of the facial feature-based subnetwork, which is $0.9136$ (see~Figure~\ref{fig:ambientaudio_vs_facial}). Therefore, it can be observed that the multimodal network, even though it consists of relatively underperformer networks, is able to give better results when compared to a model with only one modality.	

\begin{figure}[htbp]
	\centerline{
	\begin{subfigure}[b]{0.50\columnwidth}
		\includegraphics[width=\columnwidth]{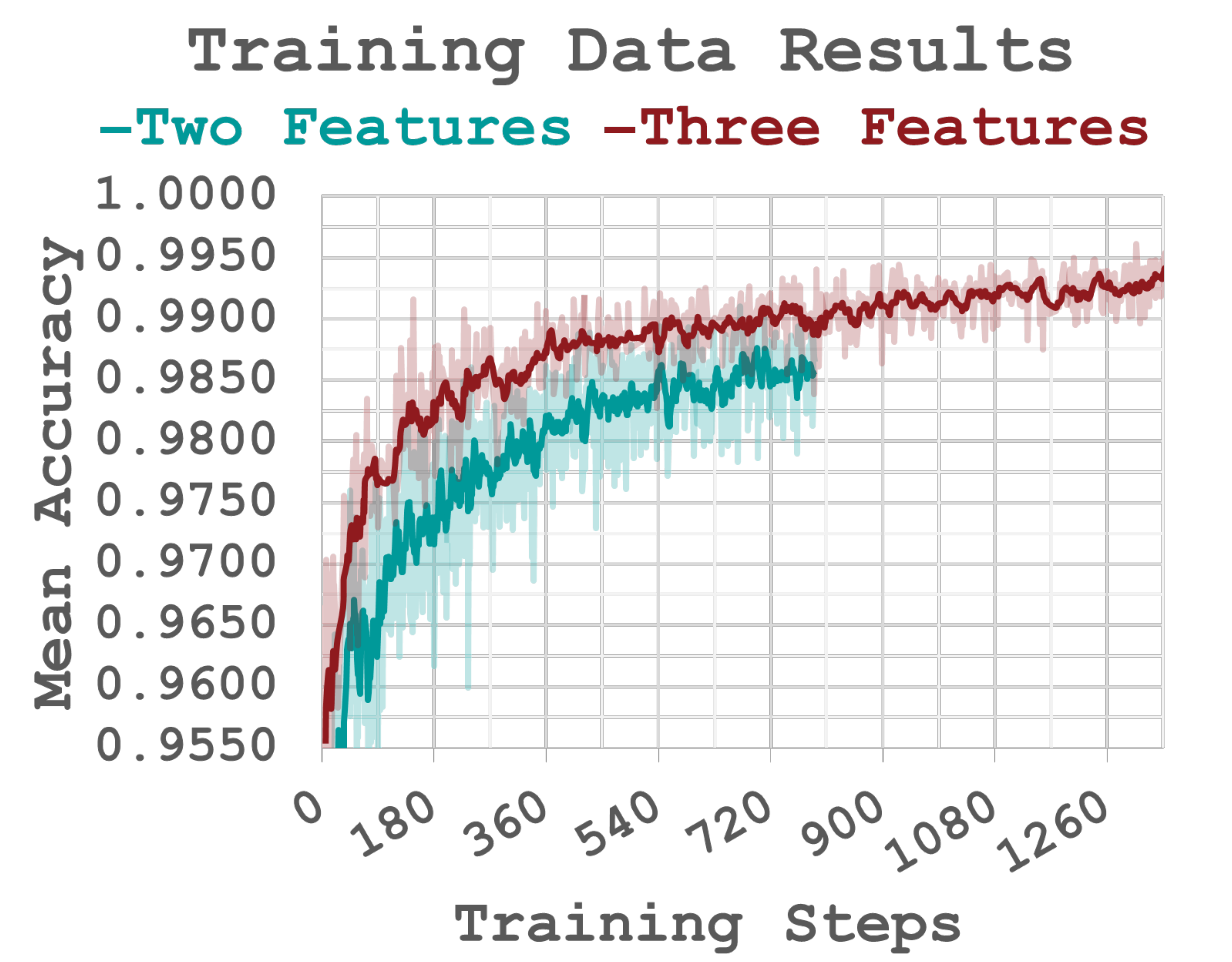}
	\end{subfigure}
	\begin{subfigure}[b]{0.50\columnwidth}
		\includegraphics[width=\columnwidth]{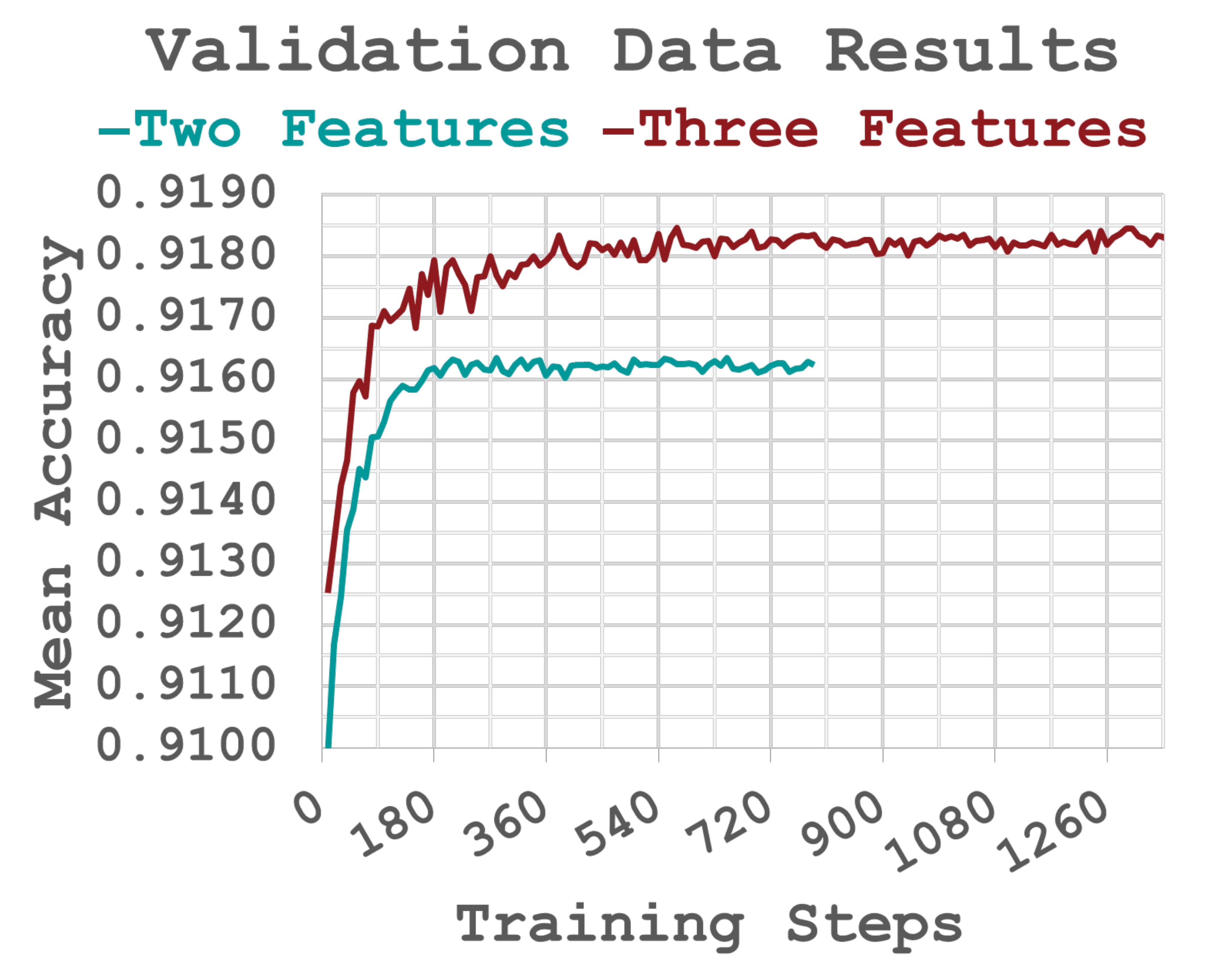}
	\end{subfigure}
	}
	\caption{The comparison of the two and three feature-based networks. The two feature-based network uses ambient and audio features and the three feature-based network uses ambient, audio, and facial features.}
	\label{fig:3feature_vs_2feature}
\end{figure}

Next, we modify the model by including the facial features in order to utilize three features. The score for the three feature-based network is $0.9185$, which is better than the two feature-based version as expected because the facial feature-based subnetwork is the best performing one (see~Figure~\ref{fig:3feature_vs_2feature}).

\begin{figure}[htbp]
	\centerline{
	\begin{subfigure}[b]{0.50\columnwidth}
		\includegraphics[width=\columnwidth]{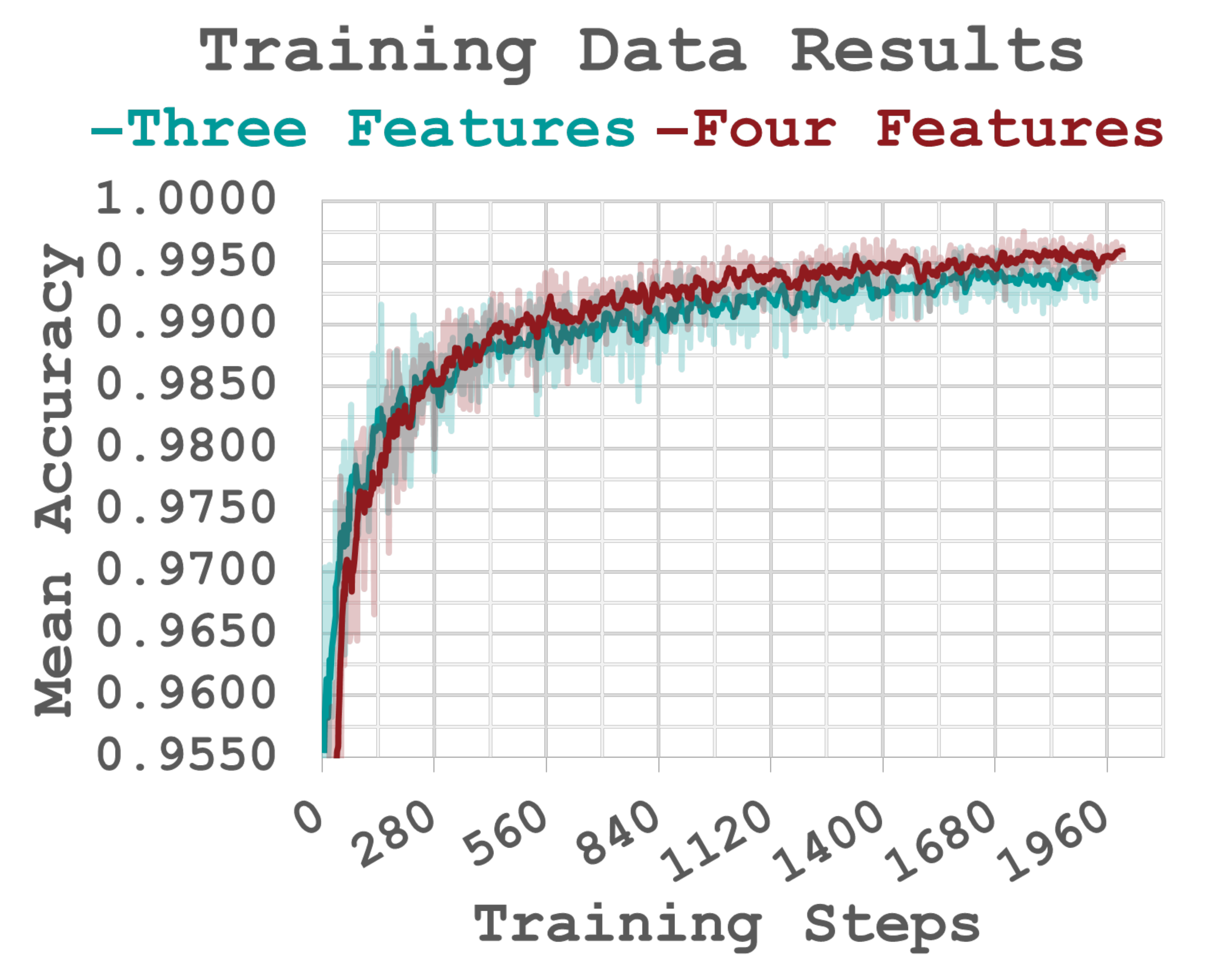}
	\end{subfigure}
	\begin{subfigure}[b]{0.50\columnwidth}
		\includegraphics[width=\columnwidth]{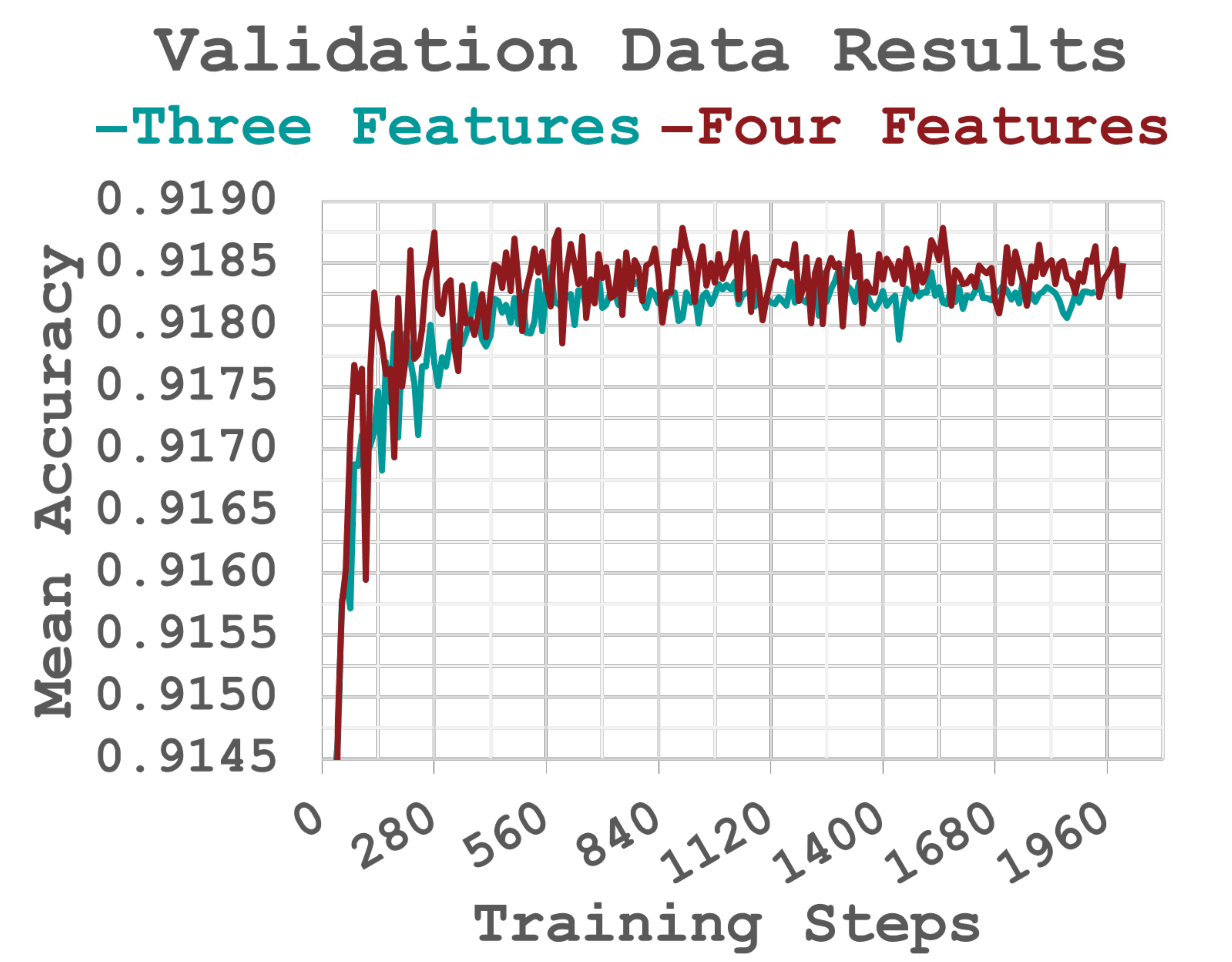}
	\end{subfigure}
	}
	\caption{The comparison of the three and four feature-based networks. The four feature-based network uses ambient, audio, facial, and transcription features.}
	\label{fig:4feature_vs_3feature}
\end{figure}

Finally, we use all features by adding the transcription input. For this four feature-based model, we made changes to the network by adding the inputs of LSTM cells to the outputs, creating a residual RNN. In the finalized model, ambient feature-based, facial feature-based and audio feature-based subnetworks have six LSTM layers, whereas transcription feature-based subnetwork has three fully connected layers. For the feature-level fusion, there are 80-dimensional features from ambient feature-based and facial feature-based subnetworks and 20-dimensional features from audio feature-based and transcription feature-based subnetworks. The final score obtained from this model is $0.9188$, which is the best score that we have obtained (see~Figure~\ref{fig:4feature_vs_3feature}). As a result, our experiments demonstrate that all modality-specific subnetworks contribute to the larger model by improving the performance.

We compare the performance of the finalized model to the state-of-the-art methods (see Table~\ref{table:results_other_methods}). We observe that the mean accuracy scores of the methods are between $0.91$ and $0.92$. In comparison to others, our approach obtains the best performance in terms of mean accuracy metric. For the individual personality traits, our method performs better than the other methods in all but one trait (openness). Apart from the validation set performances given in Table~\ref{table:results_other_methods}, NJU-LAMDA~\cite{Wei2018} yields a mean accuracy of 0.9130 and the scores of 0.9123 for openness, 0.9166 for conscientiousness, 0.9133 for extraversion, 0.9126 for agreeableness, and 0.9100 for neuroticism on the test set.

\begin{table}[h]
	\centering
	\caption{The comparison of the validation set performance of various approaches.}
	\begin{tabular}{|l|l|l|l|l|l|l|} 
		\hline
		Method 							& Mean & Open.	 & Cons.   & Extr.   & Agre. 	& Neur. \\ \hline
		DCC~\cite{guccluturk2016deep} 			& 0.9122 & 0.9117 & 0.9133 & 0.9110 & 0.9158 & 0.9091 \\ \hline
		evolgen~\cite{subramaniam2016bi} 			& 0.9134 & 0.9130 & 0.9136 & 0.9145 & 0.9157 & 0.9098 \\ \hline
		G\"{u}rp{\i}nar et al.~\cite{gurpinar2016multimodal} & 0.9147 & 0.9141 & 0.9141 & 0.9186 & 0.9143 & 0.9123 \\ \hline
		PML~\cite{eddine2017personality} 			& 0.9155 & 0.9138 & 0.9166 & 0.9175 & 0.9166 & 0.9130 \\ \hline
		BU-NKU~\cite{kaya2017multi} 				& 0.9170 & \textbf{0.9169} & 0.9166 & 0.9206 & 0.9161 & 0.9149 \\ \hline
		\textbf{The proposed method} 				& \textbf{0.9188} & 0.9166 & \textbf{0.9214} & \textbf{0.9208} & \textbf{0.9189} & \textbf{0.9162} \\ \hline
	\end{tabular}
	\label{table:results_other_methods}
\end{table}

\section{Conclusions and Future Work}
\label{section6}

We propose a novel approach for the recognition of apparent personality traits from videos. In our method, we use a multimodal neural network that consists of modality-specific CNNs to extract spatial features such as ambient features and facial expressions, and LSTM networks to integrate the temporal information of the videos. The modalities for the neural network include the face, environment, audio, and transcription features. We train the network with a two-stage training method where the modality-specific neural networks are trained to predict apparent traits independently in the first stage, and the model is fine-tuned to recognize traits accurately with a feature-level fusion of modality-specific networks in the second stage.  

First, we demonstrate that each modality-specific network is effective for personality recognition and contributes to the final model. To this end, we trained these networks separately and as a result, facial feature-based network gave the best results, followed by ambient feature-based and audio feature-based networks, and lastly, transcription feature-based network. Then, we combined the ambient feature-based and audio feature-based networks to create a simple multimodal network and compared this network to the facial feature-based network. The results showed that the multimodal approach outperforms the single modality approach. Finally, we used all modalities to obtain the finalized neural network while the improvements of each subnetwork were verified by the experiments. For the final network, we obtained best results by fine-tuning pretrained ResNet-v2-101 network with six LSTM layers for ambient feature-based and facial feature-based subnetworks, using an architecture including a pretrained VGGish network with six LSTM layers for the audio feature-based subnetwork, and using ELMo embeddings with three fully connected layers for the transcription feature-based network.

As future research directions, we envision that correlation between personality, body movements, posture, eye-gaze and emotion can be investigated to improve the performance given by the usage of ambient, face, audio, and transcription features.




\bibliographystyle{IEEEtran}
\bibliography{whole}

%

\begin{IEEEbiography}[{\includegraphics[width=1in,height=1.25in,clip,keepaspectratio]{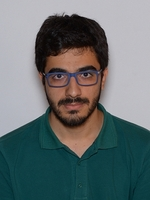}}]{S\"{u}leyman Aslan} 
received the B.S. degree in computer engineering from Bilkent University, Ankara, Turkey, in 2016. He is currently an M.S. student at the Department of Computer Engineering, Bilkent University. His research interests include computer vision, machine learning, pattern recognition, and image and video processing. He is a student member of IEEE.
\end{IEEEbiography}
\balance
\begin{IEEEbiography}[{\includegraphics[width=1in,height=1.25in,clip,keepaspectratio]{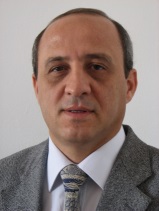}}]{U\u{g}ur G\"{u}d\"{u}kbay} (M'00-SM'05) received the B.S. degree in computer engineering from the Middle East Technical University, Ankara, Turkey, in 1987, and the M.S. and Ph.D. degrees in computer engineering and information science from Bilkent University, Ankara, in 1989 and 1994, respectively. He conducted research as a Postdoctoral Fellow at the Human Modeling and Simulation Laboratory, University of Pennsylvania, Philadelphia. Currently, he is a Professor in the Department of Computer Engineering, Bilkent University. His research interests include different aspects of computer graphics, computer vision, 3-D television, and multimedia databases. He is a senior member of ACM and IEEE.  He is serving as an Associate Editor of Computer Animation and Virtual Worlds and Signal, Image and Video Processing. 
\end{IEEEbiography}

\end{document}